\documentclass{article}

\usepackage{times}
\usepackage{graphicx} 
\usepackage{subfigure} 

\usepackage{natbib}

\usepackage{algorithm}
\usepackage{algorithmic}

\usepackage{hyperref}

\usepackage{booktabs}       
\usepackage{amsfonts}       
\usepackage{nicefrac}       
\usepackage{microtype}      
\usepackage{color} 
\usepackage{xspace}  
\usepackage[export]{adjustbox}

\usepackage{amsmath}
\usepackage[font=small,skip=10pt]{caption}

\usepackage{siunitx}

\newcommand{\todo}[1]{}
\renewcommand{\todo}[1]{{\color{red} TODO: {#1}}}

\newcommand{\note}[1]{}
\renewcommand{\note}[1]{{\color{blue} NOTE: {#1}}}

\def\ie{\emph{i.e}\ }

\usepackage[accepted]{icml2016}

\icmltitlerunning{Neural Episodic Control}

\begin{document} 

\twocolumn[
\icmltitle{Neural Episodic Control}

\icmlauthor{Alexander Pritzel}{apritzel@google.com}
\icmlauthor{Benigno Uria}{buria@google.com}
\icmlauthor{Sriram Srinivasan}{srsrinivasan@google.com}
\icmlauthor{Adri\`a Puigdom\`enech}{adriap@google.com}
\icmlauthor{Oriol Vinyals}{vinyals@google.com}
\icmlauthor{Demis Hassabis}{demishassabis@google.com}
\icmlauthor{Daan Wierstra}{wierstra@google.com}
\icmlauthor{Charles Blundell}{cblundell@google.com}
\icmladdress{DeepMind, London UK}



\icmlkeywords{reinforcement learning, deep learning}

\vskip 0.3in
]

\begin{abstract}
    Deep reinforcement learning methods attain super-human performance in a wide range of environments.
Such methods are grossly inefficient, often taking orders of magnitudes more data than humans to achieve reasonable performance.
We propose Neural Episodic Control: a deep reinforcement learning agent that is able to rapidly assimilate new experiences and act upon them. 
Our agent uses a semi-tabular representation of the value function: a buffer of past experience containing slowly changing state representations and rapidly updated estimates of the value function.
We show across a wide range of environments that our agent learns significantly faster than other state-of-the-art, general purpose deep reinforcement learning agents.
\end{abstract} 

\section{Introduction}

Deep reinforcement learning agents have achieved state-of-the-art results in a variety of complex environments~\cite{mnih2015human, mnih2016asynchronous}, often surpassing human performance~\cite{alphago}.
Although the final performance of these agents is impressive, these techniques usually require several orders of magnitude more interactions with their environment than a human in order to reach an equivalent level of expected performance.
For example, in the Atari 2600 set of environments~\cite{atari}, deep Q-networks \citep{mnih2016asynchronous} require more than 200 hours of gameplay in order to achieve scores similar to those a human player achieves after two hours \citep{lake2016building}.

The glacial learning speed of deep reinforcement learning has several plausible explanations and in this work we focus on addressing these:

1.~Stochastic gradient descent optimisation requires the use of small learning rates. Due to the global approximation nature of neural networks,  high learning rates cause catastrophic interference~\cite{catastrophic}.
Low learning rates mean that experience can only be incorporated into a neural network slowly.

2.~Environments with a sparse reward signal can be difficult for a neural network to model as there may be very few instances where the reward is non-zero.
This can be viewed as a form of class imbalance where low-reward samples outnumber high-reward samples by an unknown number.
Consequently, the neural network disproportionately underperforms at predicting larger rewards, making it difficult for an agent to take the most rewarding actions.

3.~Reward signal propagation by value-bootstrapping techniques, such as Q-learning, results in reward information being propagated one step at a time through the history of previous interactions with the environment.
This can be fairly efficient if updates happen in reverse order in which the transitions occur. However, in order to train on uncorrelated minibatches DQN-style, algorithms train on randomly selected transitions, and, in order to further stabilise training, require the use of a slowly updating \emph{target network} further slowing down reward propagation.

In this work we shall focus on addressing the three concerns listed above; we must note, however, that other recent advances in exploration \citep{osband2016deep}, hierarchical reinforcement learning \citep{vezhnevets2016strategic}
and transfer learning \citep{rusu2016progressive,fernando2017pathnet} also make substantial contributions to improving data efficiency in deep reinforcement learning over baseline agents.

In this paper we propose Neural Episodic Control (NEC), a method which tackles the limitations of deep reinforcement learning listed above and demonstrates dramatic improvements on the speed of learning for a wide range of environments.
Critically, our agent is able to rapidly latch onto highly successful strategies as soon as they are experienced, instead of waiting for many steps of optimisation (e.g., stochastic gradient descent) as is the case with DQN \citep{mnih2015human} and A3C \citep{mnih2016asynchronous}.

Our work is in part inspired by the hypothesised role of the Hippocampus in decision making \cite{ThirdWay,mfec} and also by recent work on one-shot learning \citep{vinyals2016matching} and learning to remember rare events with neural networks \cite{kaiser2016learning}.
Our agent uses a semi-tabular representation of its experience of the environment possessing several of the features of episodic memory such as long term memory, sequentiality, and context-based lookups.
The semi-tabular representation is an append-only memory that binds slow-changing keys to fast updating values and uses a context-based lookup on the keys to retrieve useful values during action selection by the agent.
Thus the agent's memory operates in much the same way that traditional table-based RL methods map from state and action to value estimates.
A unique aspect of the memory in contrast to other neural memory architectures for reinforcement learning (explained in more detail in Section~\ref{sec:methods}) is that the values retrieved from the memory can be updated much faster than the rest of the deep neural network.
This helps alleviate the typically slow weight updates of stochastic gradient descent applied to the whole network and is reminiscent of work on \emph{fast weights}
\cite{ba2016using,hinton1987using}, although the architecture we present is quite different.
Another unique aspect of the memory is that unlike other memory architectures such as LSTM and the differentiable neural computer (DNC; \citealp{graves2016hybrid}), our architecture does not try to learn when to write to memory, as this can be slow to learn and take a significant amount of time.
Instead, we elect to write all experiences to the memory, and allow it to grow very large compared to existing memory architectures (in contrast to \citet{oh2015action,graves2016hybrid} where the memory is wiped at the end of each episode). Reading from this large memory is made efficient using kd-tree based nearest neighbour \cite{Bentley1975}.

The remainder of the paper is organised as follows: in Section~\ref{sec:background} we review deep reinforcement learning, in Section~\ref{sec:methods} the Neural Episodic Control algorithm is described, in Section~\ref{sec:experiments} we report experimental results in the Atari Learning Environment, in Section~\ref{sec:related} we discuss other methods that use memory for reinforcement learning, and finally in Section~\ref{sec:discussion} we outline future work and summarise the main advantages of the NEC algorithm.

\section{Deep Reinforcement Learning}
\label{sec:background}

The action-value function of a reinforcement learning agent~\cite{sutton_barto} is defined as $Q^\pi(s, a) = \mathbb{E}_{\pi}\left[\sum_t \gamma^t r_t \mid s, a\right]$,
where $a$ is the initial action taken by the agent in the initial state $s$ and the expectation denotes that the policy $\pi$ is followed thereafter. The discount factor~$\gamma \in (0,1)$ trades off favouring short vs. long term rewards.

Deep Q-Network agents (DQN; \citealp{mnih2015human}) use Q-learning \citep{watkins1992q} to learn a value function $Q(s_t,a_t)$ to rank which action $a_t$ is best to take in each state $s_t$ at step $t$.
The agent then executes an $\epsilon$-greedy policy based upon this value function to trade-off  exploration and exploitation: with probability $\epsilon$ the agent picks an action uniformly at random, otherwise it picks the action $a_t = \arg\max_a Q(s_t, a)$.

In DQN, the action-value function $Q(s_t,a_t)$ is parameterised by a convolutional neural network that takes a 2D pixel representation of the state $s_t$ as input, and outputs a vector containing the value of each action at that state. When the agent observes a transition, DQN stores the $(s_t, a_t, r_t, s_{t+1})$ tuple in a \textit{replay buffer}, the contents of which are used for training.
This neural network is trained by minimizing the squared error between the network's output and the $Q$-learning target $y_t = r_t + \gamma \max_a \tilde{Q}(s_{t+1}, a)$, for a subset of transitions sampled at random from the replay buffer. The \textit{target network} $\tilde{Q}(s_{t+1}, a)$ is an older version of the value network that is updated periodically. The use of a target network and uncorrelated samples from the replay buffer are critical for stable training.

A number of extensions have been proposed that improve DQN.
Double DQN~\citep{van2016deep} reduces bias on the target calculation.
Prioritised Replay~\citep{schaul2015prioritized} further improves Double DQN by optimising the replay strategy.
Several authors have proposed methods of improving reward propagation and the back up mechanism of $Q$ learning 
\citep{harutyunyan2016q,munos2016safe,he2016learning} by incorporating on-policy rewards or by adding constraints to the optimisation.
Q$^*$($\lambda)$ \citep{harutyunyan2016q} and Retrace($\lambda$) \citep{munos2016safe} change the form of the Q-learning target to incorporate on-policy samples and fluidly switch between on-policy learning and off-policy learning.
\citet{munos2016safe} show that by incorporating on-policy samples allows an agent to learn faster in Atari environments, indicating that reward propagation is indeed a bottleneck to efficiency in deep reinforcement learning.

A3C~\citep{mnih2016asynchronous} is another well known deep reinforcement learning algorithm that is very different from DQN.
It is based upon a policy gradient, and learns both a policy and its associated value function, which is learned entirely on-policy (similar to the $\lambda = 1$ case of Q($\lambda$)).
Interestingly,~\citet{mnih2016asynchronous} also added an LSTM memory to the otherwise convolutional neural network architecture to give the agent a notion of memory, although this did not have significant impact on the performance on Atari games.

\section{Neural Episodic Control}
\label{sec:methods}

Our agent consists of three components: a convolutional neural network that processes pixel images $s$, a set of memory modules (one per action), and a final network that converts read-outs from the action memories into $Q(s,a)$ values.
For the convolutional neural network we use the same architecture as DQN \citep{mnih2015human}.

\subsection{Differentiable Neural Dictionary}

\begin{figure*}[ht]
    \centering
    \includegraphics[width=15cm]{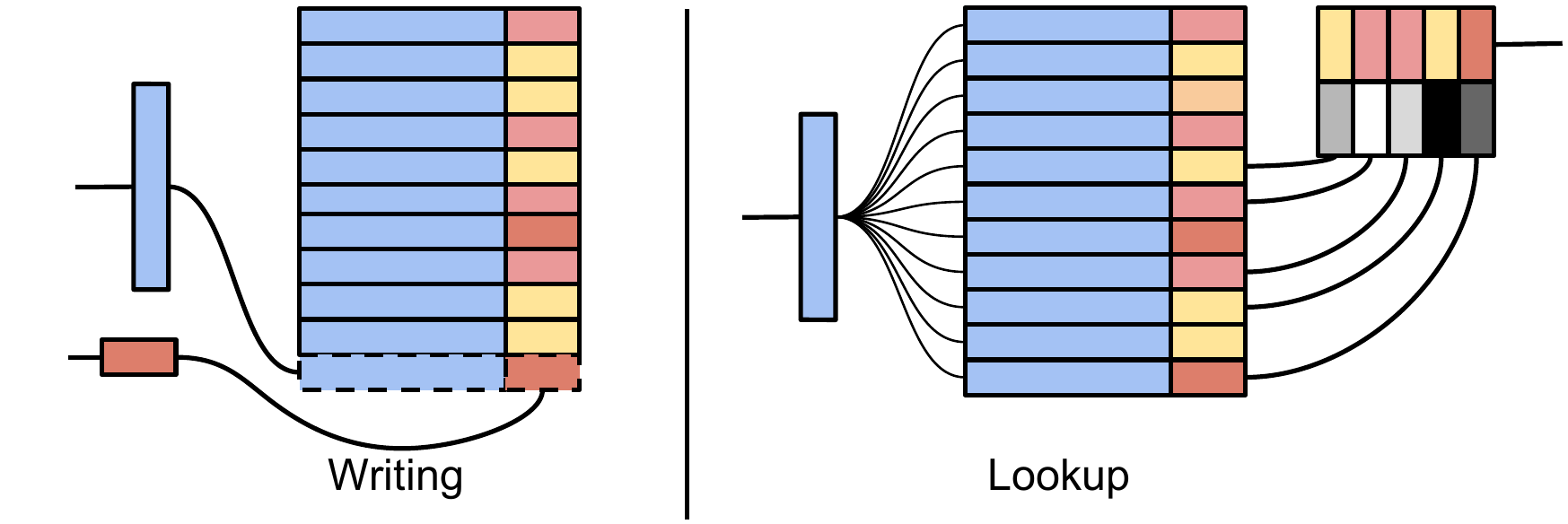}
    \caption{Illustration of operations on a Differentiable Neural Dictionary.}
    \label{fig:arch-read-write}
\end{figure*}

For each action $a\in\mathcal{A}$, NEC has a simple memory module $M_a = (K_a, V_a)$, where $K_a$ and $V_a$ are dynamically sized arrays of vectors, each containing the same number of vectors.
The memory module acts as an arbitrary association from keys to corresponding values, much like the dictionary data type found in programs.
Thus we refer to this kind of memory module as a \emph{differentiable neural dictionary} (DND).
There are two operations possible on a DND: \emph{lookup} and \emph{write}, as depicted in Figure~\ref{fig:arch-read-write}.
Performing a lookup on a DND maps a key $h$ to an output value $o$:
\begin{align}
\label{eq:outweight}
o &= \sum_i w_i v_i,
\end{align}
where $v_i$ is the $i$th element of the array $V_a$ and
\begin{equation}
    w_i = k(h, h_i)/\sum_j k(h, h_j),
    \label{eq:weighted-sum}    
\end{equation}
where $h_i$ is the $i$th element of the array $K_a$ and $k(x,y)$ is a kernel between vectors $x$ and $y$, e.g., Gaussian or inverse kernels.
Thus the output of a lookup in a DND is a weighted sum of the values in the memory, whose weights are given by normalised kernels between the lookup key and the corresponding key in memory.
To make queries into very large memories scalable we shall make two approximations in practice:
firstly, we shall limit \eqref{eq:outweight} to the top $p$-nearest neighbours (typically $p=50$).
Secondly, we use an approximate nearest neighbours algorithm to perform the lookups, based upon kd-trees \citep{Bentley1975}.

After a DND is queried, a new key-value pair is written into the memory.
The key written corresponds to the key that was looked up.
The associated value is application-specific (below we specify the update for the NEC agent).
Writes to a DND are append-only: keys and values are written to the memory by appending them onto the end of the arrays $K_a$ and $V_a$ respectively.
If a key already exists in the memory, then its corresponding value is updated, rather than being duplicated.

Note that a DND is a differentiable version of the memory module described in \citet{mfec}.
It is also a generalisation to the memory and lookup schemes described in \citep{vinyals2016matching,kaiser2016learning} for classification. 

\subsection{Agent Architecture}

Figure~\ref{fig:arch-dnd} shows a DND as part of the NEC agent for a single action, whilst Algorithm~\ref{alg:episodic} describes the general outline of the NEC algorithm.
The pixel state $s$ is processed by a convolutional neural network to produce a key $h$.
The key $h$ is then used to lookup a value from the DND, yielding weights $w_i$ in the process for each element of the memory arrays.
Finally, the output is a weighted sum of the values in the DND.
The values in the DND, in the case of an NEC agent, are the $Q$ values corresponding to the state that originally resulted in the corresponding key-value pair to be written to the memory.
Thus this architecture produces an estimate of $Q(s,a)$ for a single given action $a$.
The architecture is replicated once for each action $a$ the agent can take, with the convolutional part of the network shared among each separate DND $M_a$.
The NEC agent acts by taking the action with the highest $Q$-value estimate at each time step.
In practice, we use $\epsilon$-greedy policy during training with a low $\epsilon$.

\begin{algorithm}[h]
    \caption{Neural Episodic Control
		\label{alg:episodic}}
	\begin{algorithmic}
	    \STATE $\mathcal{D}$: replay memory.
        \STATE $M_a$: a DND for each action $a$.
        \STATE $N$: horizon for $N$-step $Q$ estimate.
		\FOR{each episode}
            \FOR{$t = 1, 2, \dots, T$}
                \STATE Receive observation $s_t$ from environment with embedding $h$.
                \STATE Estimate $Q(s_t, a)$ for each action $a$ via \eqref{eq:outweight} from $M_a$
                \STATE $a_t \leftarrow$ $\epsilon$-greedy policy based on $Q(s_t,a)$
                \STATE Take action $a_t$, receive reward $r_{t+1}$
                \STATE Append $(h,Q^{(N)}(s_t,a_t))$ to $M_{a_t}$.
                \STATE Append $(s_t, a_t, Q^{(N)}(s_t,a_t))$ to $\mathcal{D}$.
                \STATE Train on a random minibatch from $\mathcal{D}$.
		    \ENDFOR
		\ENDFOR
	\end{algorithmic}
	\label{alg:nec}
\end{algorithm}

\subsection{Adding $(s,a)$ pairs to memory}

As an NEC agent acts, it continually adds new key-value pairs to its memory.
Keys are appended to the memory of the corresponding action, taking the value of the query key $h$ encoded by the convolutional neural network.
We now turn to the question of an appropriate corresponding value.
In \citet{mfec}, Monte Carlo returns were written to memory.
We found that a mixture of Monte Carlo returns (on-policy) and off-policy backups worked better and so for NEC we elect to use $N$-step $Q$-learning as in \citet{mnih2016asynchronous} (see also \citealp{watkins1989learning,peng1996incremental}).
This adds the following $N$ on-policy rewards and bootstraps the sum of discounted rewards for the rest of the trajectory, off-policy.
The $N$-step $Q$-value estimate is then
\begin{equation}
    Q^{(N)}(s_t,a) = \sum_{j=0}^{N-1} \gamma^j r_{t+j} + \gamma^N \max_{a'} Q(s_{t+N}, a')\ .
    \label{eq:nstepq}
\end{equation}
The bootstrap term of \eqref{eq:nstepq}, $\max_{a'} Q(s_{t+N}, a')$ is found by querying all memories $M_a$ for each action $a$ and taking the highest estimated $Q$-value returned.
Note that the earliest such values can be added to memory is $N$ steps after a particular $(s,a)$ pair occurs.

When a state-action value is already present in a DND (\ie~the exact same key $h$ is already in $K_a$), the corresponding value present in $V_a$, $Q_i$, is updated in the same way as the classic tabular $Q$-learning algorithm:
\begin{equation}
Q_i \leftarrow Q_i + \alpha (Q^{(N)}(s,a) - Q_i)\ . 
\label{eq:fast-updates}
\end{equation}
where $\alpha$ is the learning rate of the $Q$ update.
If the state is not already present $Q^{(N)}(s_t,a)$ is appended to $V_a$ and $h$ is appended to $K_a$.
Note that our agent learns the value function in much the same way that a classic tabular $Q$-learning agent does, except that the $Q$-table grows with time.
We found that $\alpha$ could take on a high value, allowing repeatedly visited states with a stable representation to rapidly update their value function estimate.
Additionally, batching up memory updates (e.g., at the end of the episode) helps with computational performance.
We overwrite the item that has least recently shown up as a neighbour when we reach the memory's maximum capacity.

\subsection{Learning}

\begin{figure*}[h]
    \centering
    \includegraphics[width=15cm]{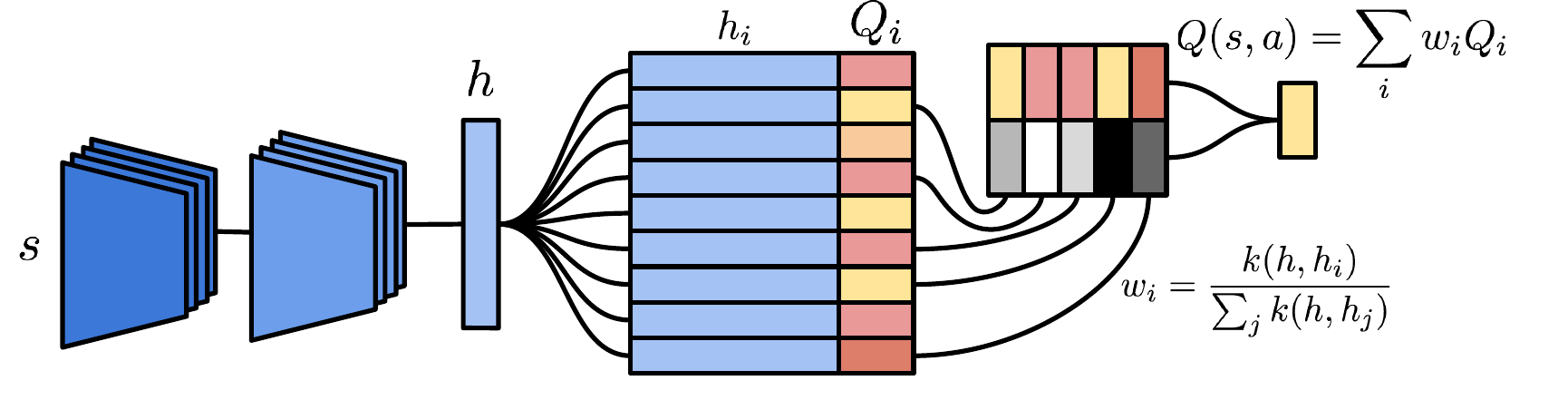}
    \caption{Architecture of episodic memory module for a single action $a$. Pixels representing the current state enter through a convolutional neural network on the bottom left and
    an estimate of $Q(s,a)$ exits top right. Gradients flow through the entire architecture.}
    \label{fig:arch-dnd}
\end{figure*}

Agent parameters are updated by minimising the $L2$ loss between the predicted $Q$ value for a given action and the $Q^{(N)}$ estimate on randomly sampled mini-batches from a replay buffer.
In particular, we store tuples $(s_t, a_t,R_{t})$
 in the replay buffer, where $N$ is the horizon of the $N$-step Q rule, and $R_{t} = Q^{(N)}(s_{t},a)$ plays the role of the target network seen in DQN (our replay buffer is significantly smaller than DQN's).
These $(s_t, a_t, R_{t})$-tuples are then sampled uniformly at random to form minibatches for training.
Note that the architecture in Figure~\ref{fig:arch-dnd} is entirely differentiable and so we can minimize this loss by gradient descent.
Backpropagation updates the the weights and biases of the convolutional embedding network and the keys and values of each action-specific memory using gradients of this loss, using a lower learning rate than is used for updating pairs after queries ($\alpha$).

\section{Experiments}
\label{sec:experiments}

We investigated whether neural episodic control allows for more data efficient learning in practice in complex domains.
As a problem domain we chose the Atari Learning Environment(ALE;~\citealp{atari}). We tested our method on the 57 Atari games used by \citet{PrioritizedReplay}, which form an interesting set of tasks as they contain diverse challenges such as sparse rewards and vastly different magnitudes of scores across games. 
Most common algorithms applied in these domains, such as variants of DQN and A3C, require in the thousands of hours of in-game time, i.e. they are data inefficient.

We consider 5 variants of A3C and DQN as baselines as well as MFEC \citep{mfec}.
We compare to the basic implementations of A3C~\cite{mnih2016asynchronous} and DQN~\cite{mnih2015human}. We also compare to two algorithms incorporating $\lambda$ returns~\cite{sutton1988} aiming at more data efficiency by faster propagation of credit assignments, namely $Q^*(\lambda)$~\citep{harutyunyan2016q} and $\text{Retrace}(\lambda)$~\citep{munos2016safe}. 
We also compare to DQN with Prioritised Replay, which improves data efficiency by replaying more salient transitions more frequently.
We did not directly compare to DRQN~\citep{hausknecht2015deep} nor FRMQN~\citep{oh2016control} as results were not available for all Atari games.
Note that in the case of DRQN, reported performance is lower than that of Prioritised Replay.

All algorithms were trained using discount rate $\gamma=0.99$, except MFEC that uses $\gamma=1$. In our implementation of MFEC we used random projections as an embedding function, since in the original publication it obtained better performance on the Atari games tested. 

In terms of hyperparameters for NEC,  we chose the same convolutional architecture as DQN,
and store up to $5\times 10^5$ memories per action. We used the RMSProp algorithm~\cite{tieleman2012lecture} for gradient descent training.
We apply the same preprocessing steps as \cite{mnih2015human}, including repeating each action four times.
For the $N$-step $Q$ estimates we picked a horizon of $N=100$.
Our replay buffer stores the only last $10^5$ states (as opposed to $10^6$ for DQN) observed and their $N$-step $Q$ estimates. We do one replay update for every 16 observed frames with a minibatch of size 32.
We set the number of nearest neighbours $p=50$ in all our experiments. For the kernel function we chose a function that interpolates between the mean for short distances and weighted inverse distance for large distances, more precisely:
\hspace*{-0.1cm}
\begin{equation}
    k(h, h_i) = \frac{1}{\|h-h_i\|_2^2 + \delta}.
\end{equation}
Intuitively, when all neighbours are far away we want to avoid putting all weight onto one data point. A Gaussian kernel, for example, would exponentially suppress all neighbours except for the closest one. The kernel we chose has the advantage of having heavy tails. This makes the algorithm more robust and we found it to be less sensitive to kernel hyperparameters. We set $\delta=10^{-3}$.

In order to tune the remaining hyperparameters (SGD learning-rate, fast-update learning-rate $\alpha$ in Equation~\ref{eq:fast-updates}, dimensionality of the embeddings, $Q^{(N)}$ in Equation~\ref{eq:nstepq}, and $\epsilon$-greedy exploration-rate) we ran a hyperparameter sweep on six games: Beam Rider, Breakout, Pong, Q*Bert, Seaquest and Space Invaders.
We picked the hyperparameter values that performed best on the median for this subset of games (a common cross validation procedure described by~\citet{bellemare2012arcade}, and adhered to by~\citet{mnih2015human}).

\begin{table*}
\centering
\begin{tabular}{llllllll}
\hline
Frames & Nature DQN   & Q$^*(\lambda)$   & Retrace$(\lambda)$   & Prioritised Replay   & A3C   & NEC            & MFEC   \\
\hline
 1M  & -0.7\%        & -0.8\%            & -0.4\%                & -2.4\%                & 0.4\%  & \textbf{16.7\%} & 12.8\%  \\
 2M  & 0.0\%         & 0.1\%             & 0.2\%                 & 0.0\%                 & 0.9\%  & \textbf{27.8\%} & 16.7\%  \\
 4M  & 2.4\%         & 1.8\%             & 3.3\%                 & 2.7\%                 & 1.9\%  & \textbf{36.0\%} & 26.6\%  \\
 10M & 15.7\%        & 13.0\%            & 17.3\%                & 22.4\%                & 3.6\%  & \textbf{54.6\%} & 45.4\%  \\
 20M & 26.8\%        & 26.9\%            & 30.4\%                & 38.6\%                & 7.9\%  & \textbf{72.0\%} & 55.9\%  \\
 40M & 52.7\%        & 59.6\%            & 60.5\%                & \textbf{89.0\%}       & 18.4\% & 83.3\%          & 61.9\%  \\
\hline
\end{tabular}
\caption{Median across games of human-normalised scores for several algorithms at different points in training}
\label{tab:median-at}
\end{table*}
\begin{table*}
\centering
\begin{tabular}{llllllll}
\hline
Frames & Nature DQN   & Q$^*(\lambda)$   & Retrace$(\lambda)$   & Prioritised Replay   & A3C    & NEC            & MFEC   \\
\hline
 1M  & -10.5\%       & -11.7\%           & -10.5\%               & -14.4\%               & 5.2\%   & \textbf{45.6\%} & 28.4\%  \\
 2M  & -5.8\%        & -7.5\%            & -5.4\%                & -5.4\%                & 8.0\%   & \textbf{58.3\%} & 39.4\%  \\
 4M  & 8.8\%         & 6.2\%             & 6.2\%                 & 10.2\%                & 11.8\%  & \textbf{73.3\%} & 53.4\%  \\
 10M & 51.3\%        & 46.3\%            & 52.7\%                & 71.5\%                & 22.3\%  & \textbf{99.8\%} & 85.0\%  \\
 20M & 94.5\%        & 135.4\%           & \textbf{273.7\%}      & 165.2\%               & 59.7\%  & 121.5\%         & 113.6\% \\
 40M & 151.2\%       & \textbf{440.9\%}  & 386.5\%               & 332.3\%               & 255.4\% & 144.8\%         & 142.2\% \\
\hline
\end{tabular}
\caption{Mean human-normalised scores for several algorithms at different points in training}
\label{tab:mean-at}
\end{table*}

Data efficiency results are summarised in Table~\ref{tab:median-at}. In the small data regime (less than 20 million frames) NEC clearly outperforms all other algorithms. The difference is especially pronounced before 5 million frames have been observed. Only at 40 million frames does DQN with Prioritised Replay outperform NEC on average; note that this corresponds to 185 hours of gameplay.

\begin{figure}[h]
\centering
\includegraphics[width=9cm]{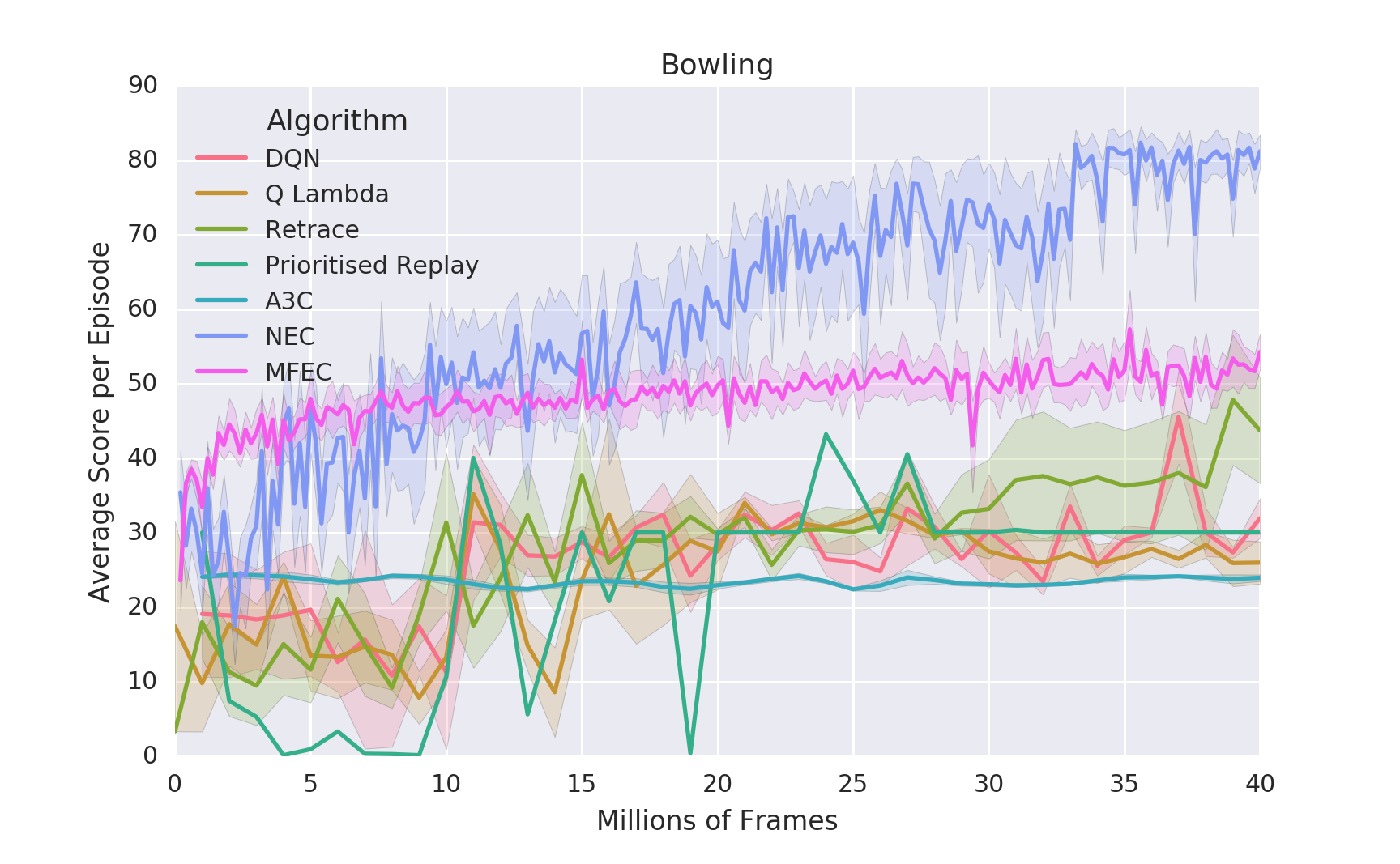}
\caption{Learning curve on Bowling.}
\label{fig:curves-bowling}
\end{figure}

\begin{figure}[h]
\centering
\includegraphics[width=9cm]{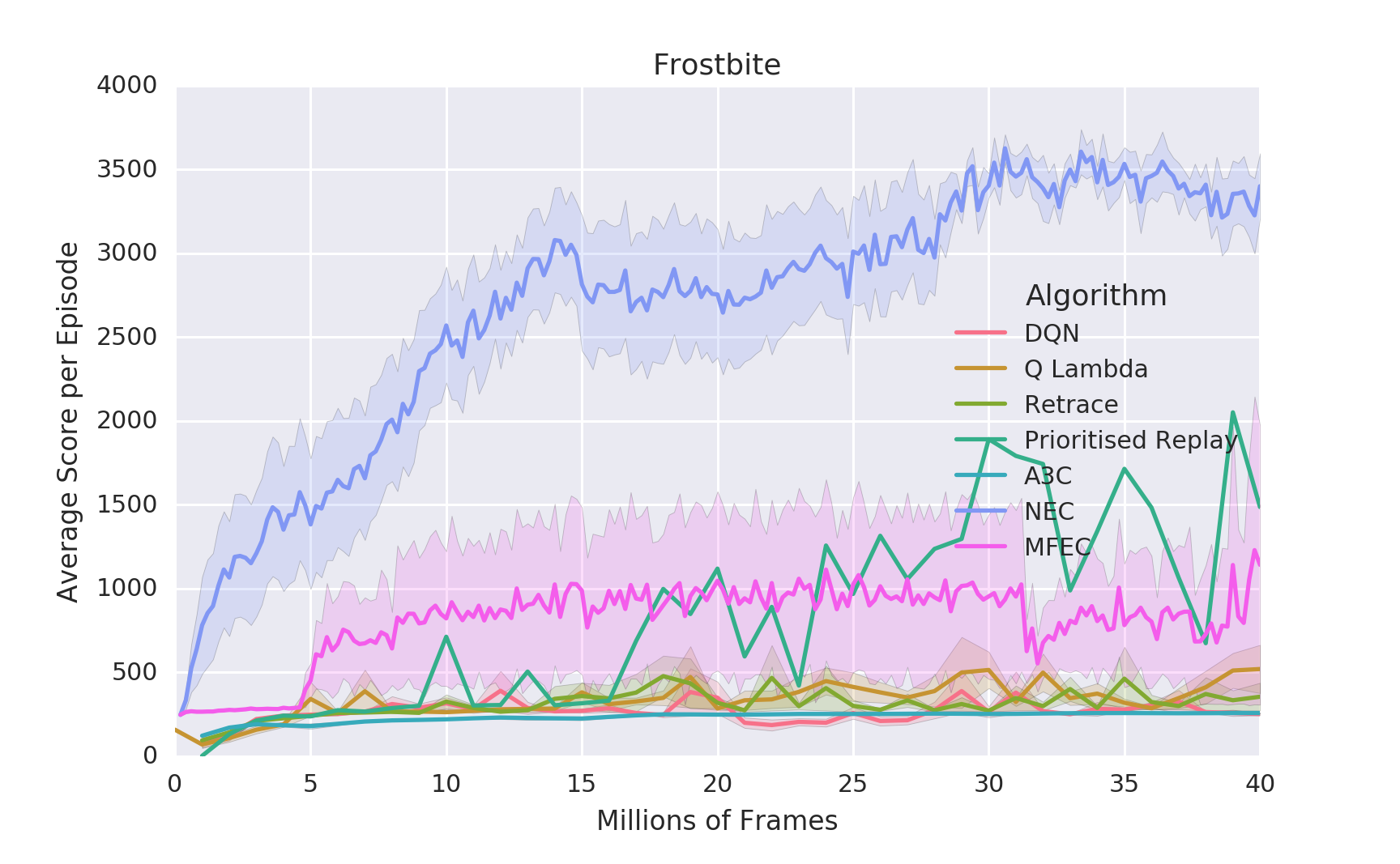}
\caption{Learning curve on Frostbite.}
\label{fig:curves-frostbite}
\end{figure}

\begin{figure}[h]
\centering
\includegraphics[width=9cm]{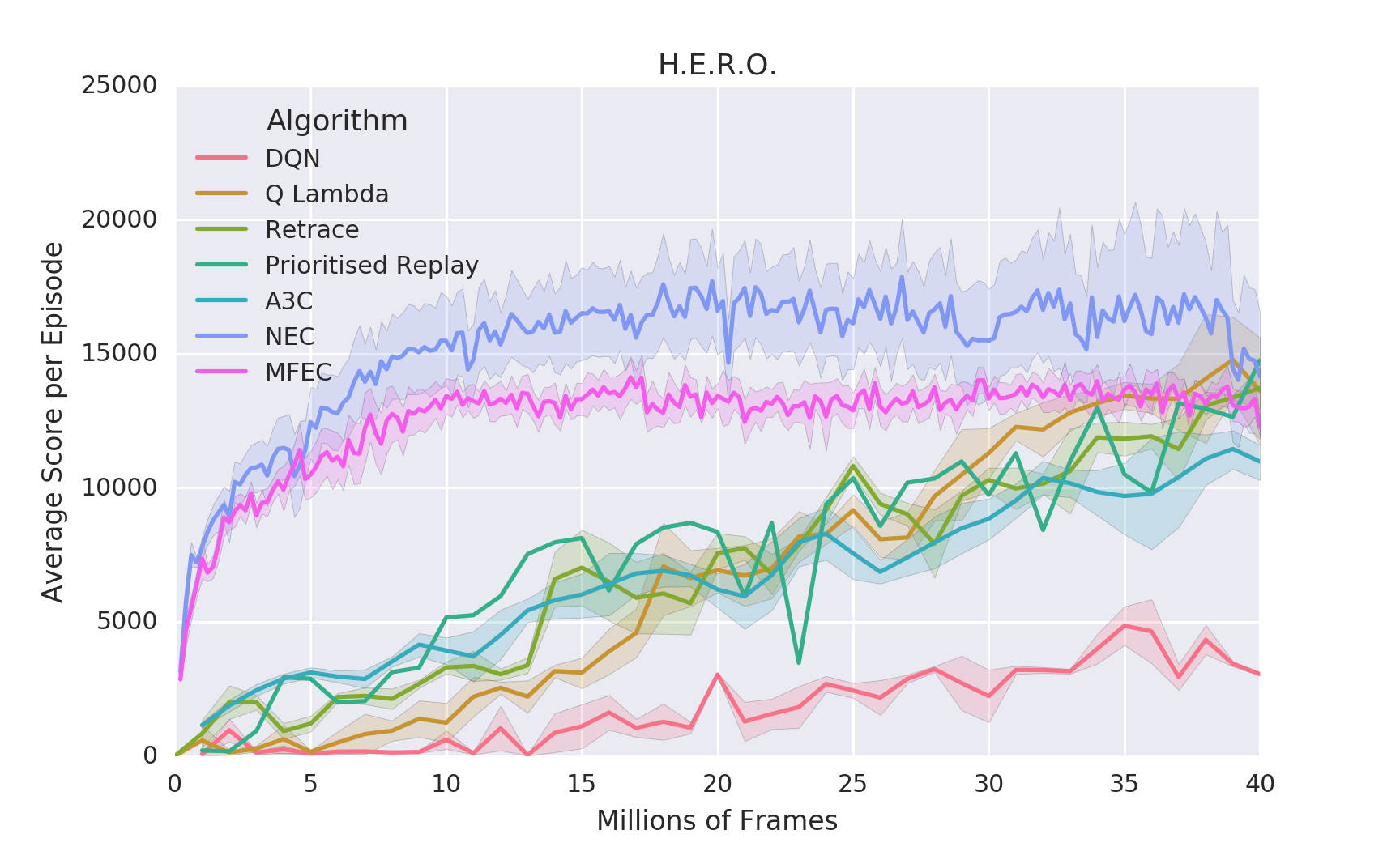}
\caption{Learning curve on H.E.R.O.}
\label{fig:curves-hero}
\end{figure}

\begin{figure}[h]
\centering
\includegraphics[width=9cm]{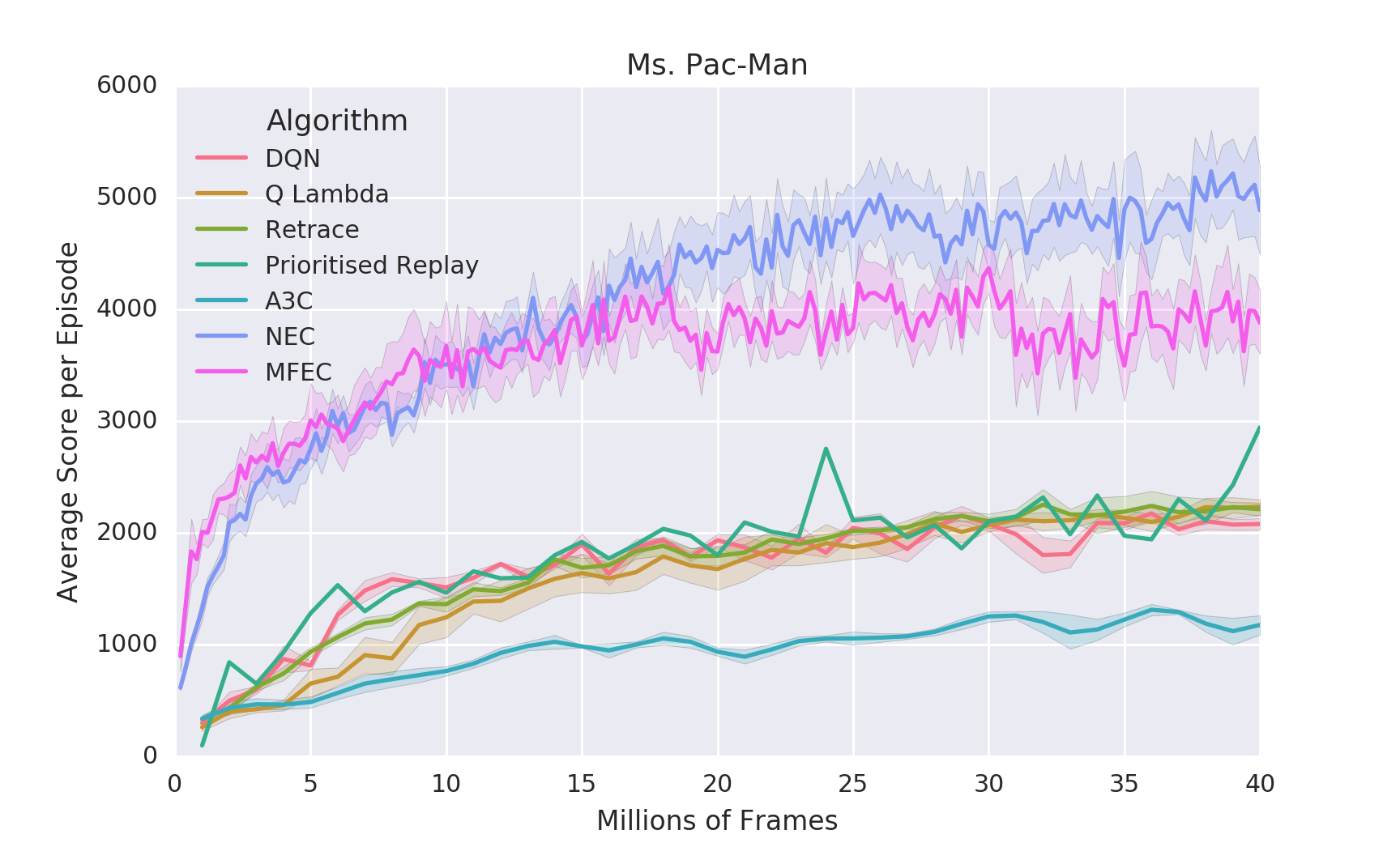}
\caption{Learning curve on Ms. Pac-Man.}
\label{fig:curves-mspacman}
\end{figure}

\begin{figure}[h]
\centering
\includegraphics[width=9cm]{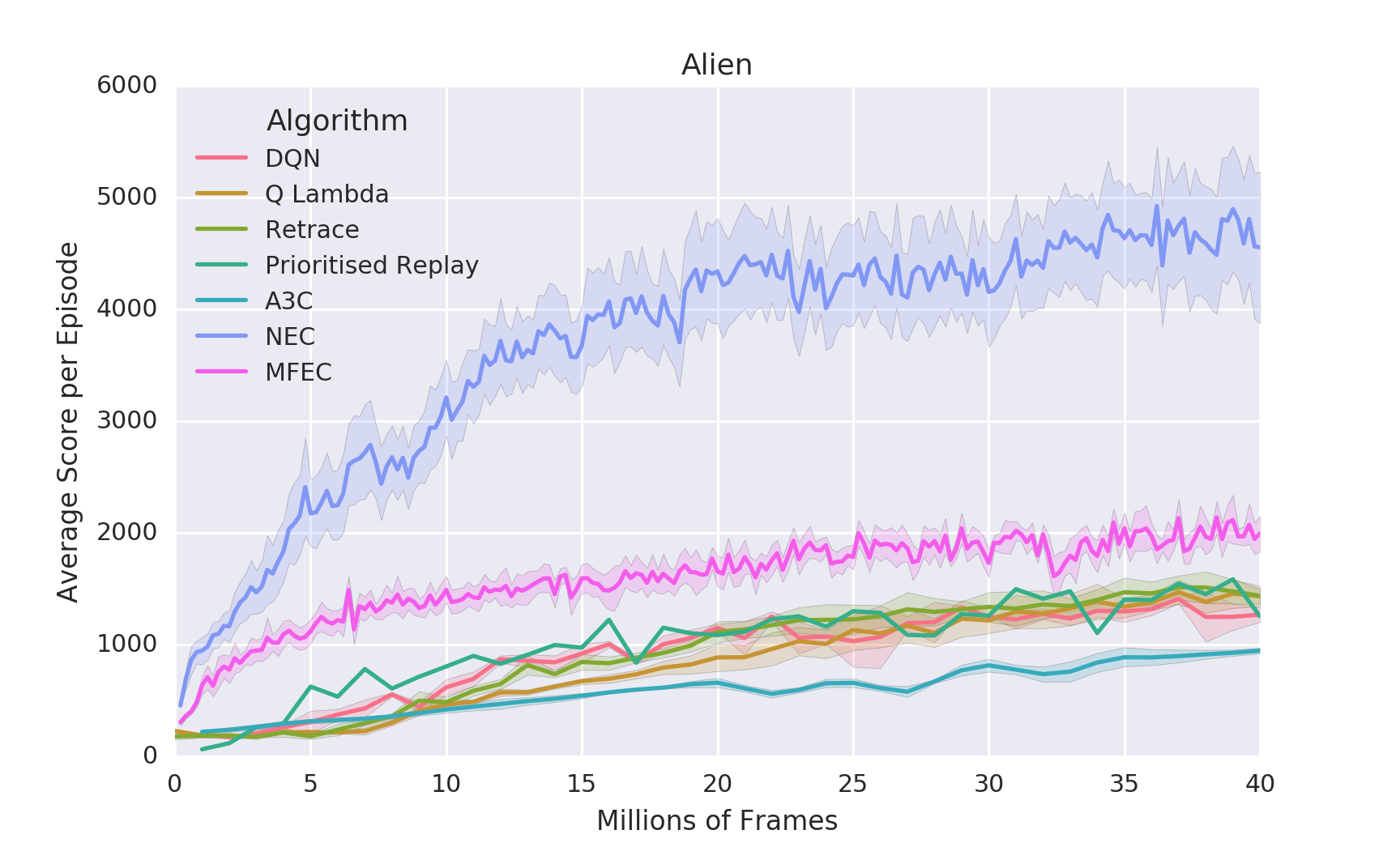}
\caption{Learning curve on Alien.}
\label{fig:curves-alien}
\end{figure}

In order to provide a more detailed picture of NEC's performance, Figures~\ref{fig:curves-bowling} to~\ref{fig:curves-alien} show learning curves on 6 games (Alien, Bowling, Boxing, Frostbite, HERO, Ms. Pac-Man, Pong), where several stereotypical cases of NEC's performance can be observed. All learning curves show the average performance over $5$ different initial random seeds. We evaluate MFEC and NEC every $200.000$ frames, and the other algorithms are evaluated every million steps.

Across most games, NEC is significantly faster at learning in the initial phase~(see also Table~\ref{tab:median-at}), only comparable to MFEC, which also uses an episodic-like $Q$-function.

NEC also outperforms MFEC on average~(see Table~\ref{tab:mean-at}). 
In contrast with MFEC, NEC uses the reward signal to learn an embedding adequate for value interpolation. 
This difference is especially significant in games where a few pixels determine the value of each action. 
The simpler version of MFEC uses an approximation to $L2$ distances in pixel-space by means of random projections, and cannot focus on the small but most relevant details. 
Another version of MFEC calculated distances on the latent representation of a variational autoencoder~\cite{kingma2013auto} trained to model frames. 
This latent representation does not depend on rewards and will be subject to irrelevant details like, for example, the display of the current score.

A3C, DQN and related algorithms require rewards to be clipped to the range $[-1, 1]$ for training stability\footnote{See \textit{Pop--Art}~\cite{vanHasselt2016} for a DQN-like algorithm that does not require reward-clipping. 
NEC also outperforms \textit{Pop--Art}. }\cite{mnih2015human}. 
NEC and MFEC do not require reward clipping, which results in qualitative changes in behaviour and better performance relative to other algorithms on games requiring clipping (Bowling, Frostbite, H.E.R.O., Ms. Pac-Man, Alien out of the seven shown). 

Alien and Ms. Pac-Man both involve controlling a character, where there is an easy way to collect small rewards by collecting items of which there are plenty, while avoiding enemies, which are invulnerable to the agent. On the other hand the agent can pick up a special item making enemies vulnerable, allowing the agent to attack them and get significantly larger rewards than from collecting the small rewards. Agents trained using existing parametric methods tend to show little interest in this as clipping implies there is no difference between large and small rewards. Therefore, as NEC does not need reward clipping, it can strongly outperform other algorithms, since NEC is maximising the non-clipped score (the true score). This can also be seen when observing the agents play: parametric methods will tend to collect small rewards, while NEC will try to actively make the enemies vulnerable and attack them to get large rewards.

\begin{figure}[h]
\centering
\includegraphics[width=9cm]{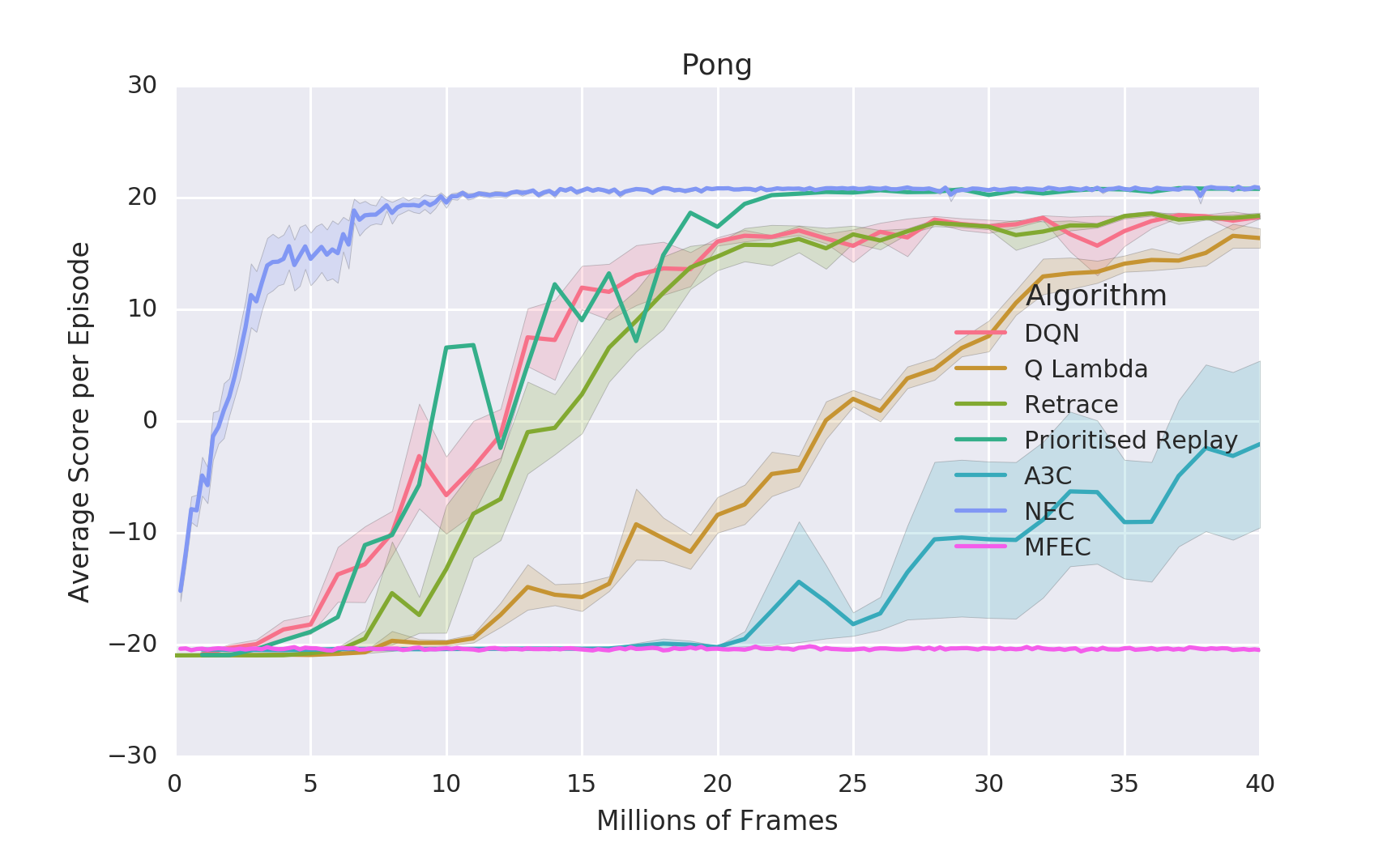}
\caption{Learning curve on Pong.}
\label{fig:curves-pong}
\end{figure}

\begin{figure}[h]
\centering
\includegraphics[width=9cm]{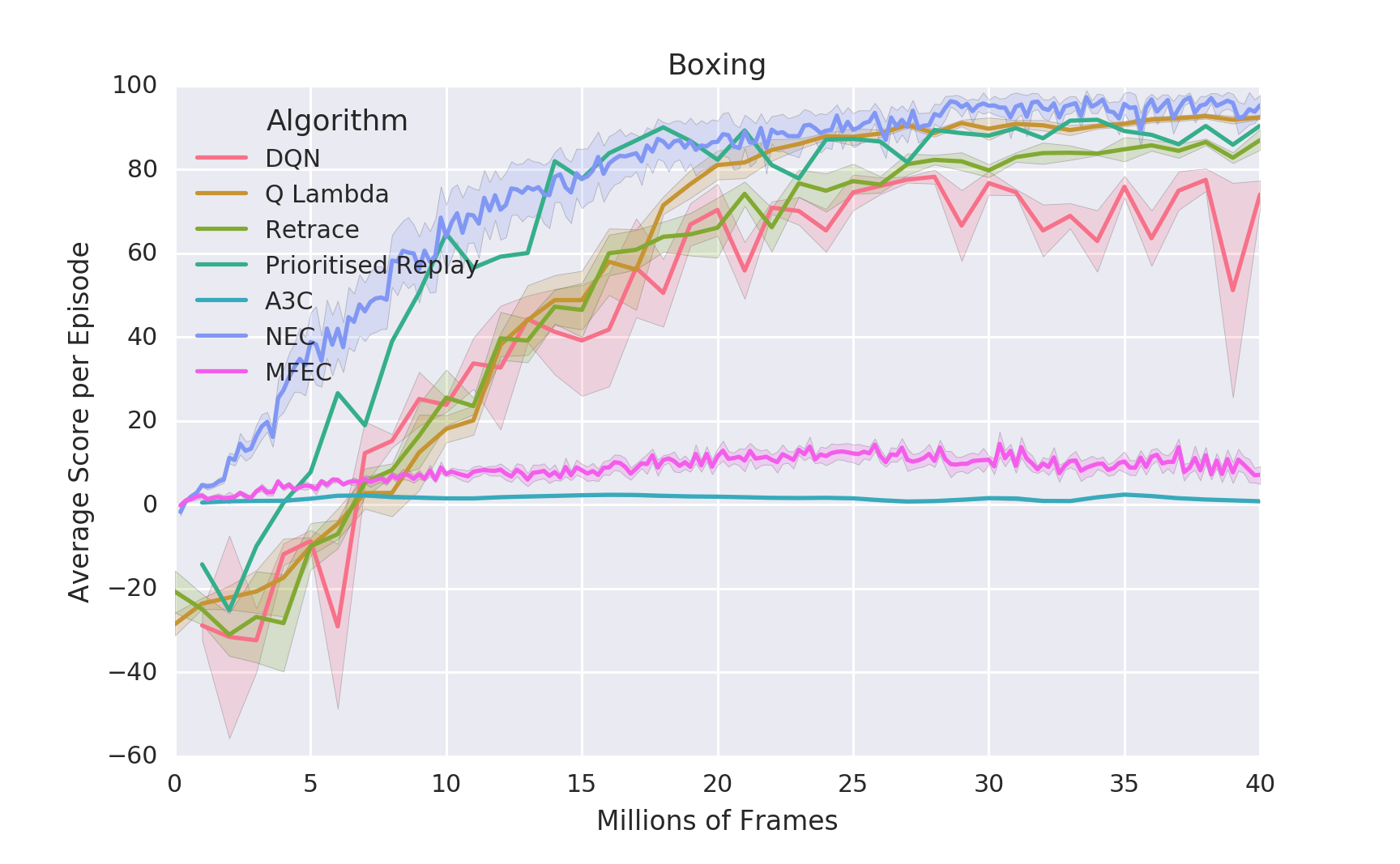}
\caption{Learning curve on Boxing.}
\label{fig:curves-boxing}
\end{figure}

NEC also outperforms the other algorithms on Pong and Boxing where reward clipping does not affect any of the algorithms as all original rewards are in the range $[-1, 1]$; as can be expected, NEC does not outperform others in terms of maximally achieved score, but it is vastly more data efficient.

\begin{figure}[h]
\centering
\includegraphics[scale=0.5]{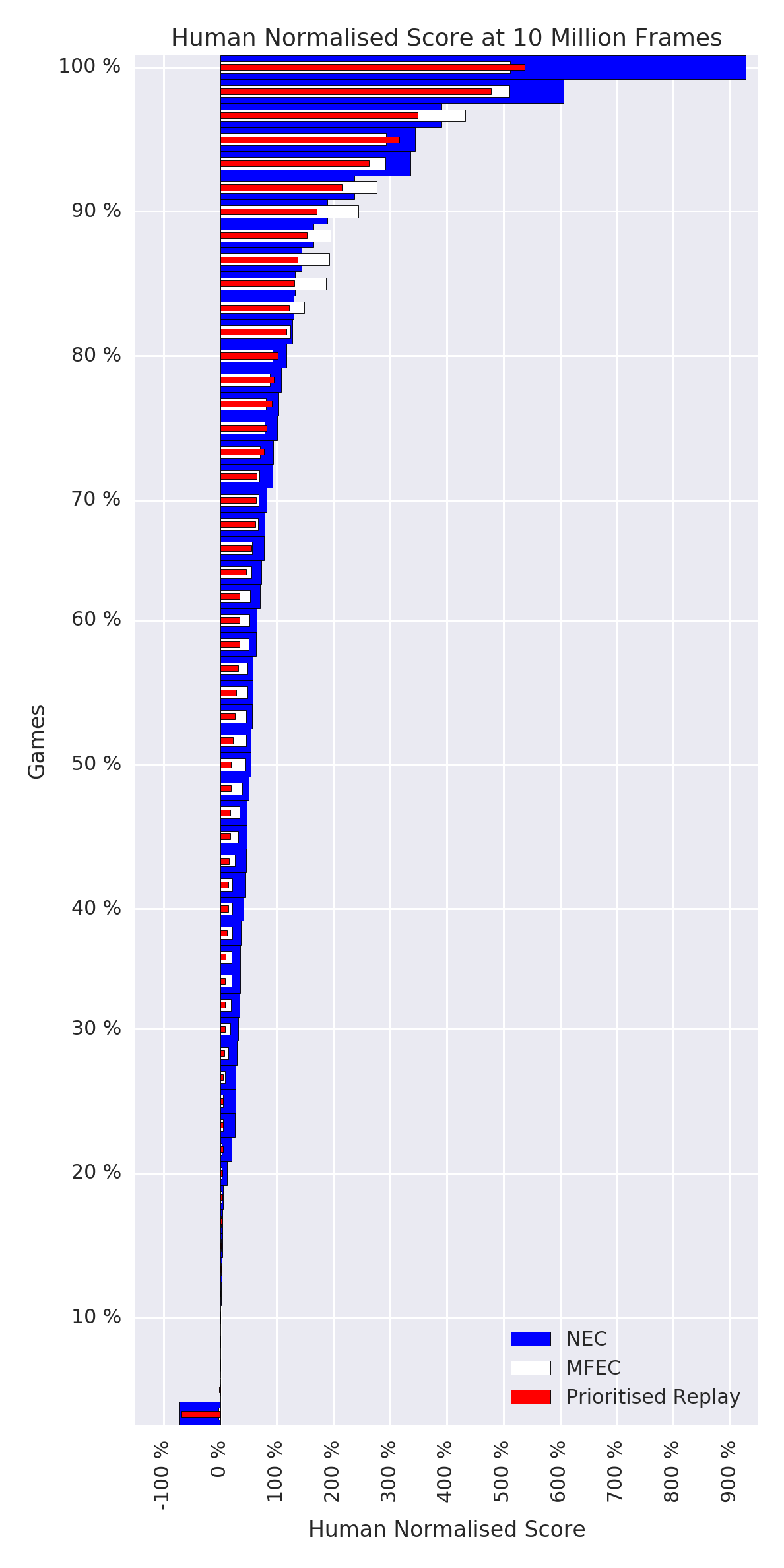}
\caption{Human-normalised scores of games, independently ranked per algorithm; labels on y-axis denote quantiles.}
\label{fig:bar-chart}
\end{figure}

In Figure~\ref{fig:bar-chart} we show a chart of human-normalised scores across all 57 Atari games at 10 million frames comparing to Prioritised Replay and MFEC. 
We rank the games independently for each algorithm, and on the y-axis the deciles are shown.

We can see that NEC gets to a human level performance in about $25\%$ of the games within $10$ million frames.
As we can see NEC outperforms MFEC and Prioritised Replay.

\section{Related work}
\label{sec:related}

There has been much recent work on memory architectures for neural networks (LSTM; \citealp{hochreiter1997long}), DNC \citep{graves2016hybrid}, memory networks \citep{sukhbaatar2015end,miller2016key}). Recurrent neural network representations of memory (LSTMs and DNCs) are trained by truncated backpropagation through time, and are subject to the same slow learning of non-recurrent neural networks.

Some of these models have been adapted to their use in RL agents (LSTMs; \citealp{bakker2003robot,hausknecht2015deep}), DNCs \citep{graves2016hybrid}, memory networks \citep{oh2016control}. However, the contents of these memories is typically reset at the beginning of every episode. This is appropriate when the goal of the memory is tracking previous observations in order to maximise rewards in partially observable or non-Markovian environments. Therefore, these implementations can be thought of as a type of working memory, and solve a different problem than the one addressed in this work.

RNNs can learn to quickly write highly rewarding states into memory and may even be able to learn entire reinforcement learning algorithms \citep{wang2016learning,duan2016rl}. However, doing so can take an arbitrarily long time and the learning time likely scales strongly with the complexity of the task.

The work of \citet{oh2016control} is also reminiscent of the ideas presented here. They introduced (FR)MQN, an adaptation of memory networks used in the top layers of a $Q$-network.

\citet{kaiser2016learning} introduced a differentiable layer of key-value pairs that can be plugged into a neural network. This layer uses cosine similarity to calculate a weighted average of the values associated with the $k$ most similar memories. Their use of a moving average update rule is reminiscent of the one presented in Section~\ref{sec:methods}. The authors reported results on a set of supervised tasks, however they did not consider applications to reinforcement learning.
Other deep RL methods keep a history of previous experience. 
Indeed, DQN itself has an elementary form of memory: the replay buffer central to its stable training can be viewed as a memory that is frequently replayed to distil the contents into DQN's value network.
\citet{kumaran2016learning} suggest that training on replayed experiences from the replay buffer in DQN is similar to the replay of experiences from episodic memory during sleep in animals.
DQN's replay buffer differs from most other work on memory for deep reinforcement learning in its sheer scale: it is common for DQN's replay buffer to hold millions of $(s,a,r,s')$ tuples.
The use of local regression techniques for $Q$-function approximation has been suggested before: 
\citet{santamaria} proposed the use of k-nearest-neighbours regression with a heuristic for adding memories based on the distance to previous memories.
\citet{barycentric} proposed barycentric interpolators to model the value function and proved their convergence to the optimal value function under mild conditions, but no empirical results were presented. 
\citet{riedmiller2005} also suggested the use of local regression, under the paradigm of case-based-reasoning that included heuristics for the deletion of stored cases. 
\citet[MFEC]{mfec} recently used local regression for $Q$-function estimation using the mean of the k-nearest neighbours, except in the case of an exact match of the query point, in which case the stored value was returned. They also propose the use of the latent variable obtained from a variational autoencoder~\cite{rezende2014stochastic} as an embedding space, but showed random projections often obtained better results. 
In contrast with the ideas presented here, none of the local-regression work aforementioned uses the reward signal to learn an embedding space of covariates in which to perform the local-regression. 
We learn this embedding space using temporal-difference learning; a crucial difference, as we showed in the experimental comparison to MFEC.
\section{Discussion}
\label{sec:discussion}

We have proposed Neural Episodic Control (NEC): a deep reinforcement learning agent that learns significantly faster than other baseline agents on a wide range of Atari 2600 games.
At the core of NEC is a memory structure: a Differentiable Neural Dictionary (DND), one for each potential action.
NEC inserts recent state representations paired with corresponding value functions into the appropriate DND.

Our experiments show that NEC requires an order of magnitude fewer interactions with the environment than agents previously proposed for data efficiency, such as Prioritised Replay \citep{schaul2015prioritized} and Retrace($\lambda$) \citep{munos2016safe}.
We speculate that NEC learns faster through a combination of three features of the agent:
the memory architecture (DND), the use of $N$-step $Q$ estimates, and a state representation provided by a convolutional neural network.

The memory architecture, DND, rapidly integrates recent experience---state representations and corresponding value estimates---allowing this information to be rapidly integrated into future behaviour.
Such memories persist across many episodes, and we use a fast approximate nearest neighbour algorithm (kd-trees) to ensure that such memories can be efficiently accessed.
Estimating $Q$-values by using the $N$-step $Q$ value function interpolates between Monte Carlo value estimates and backed up off-policy estimates.
Monte Carlo value estimates reflect the rewards an agent is actually receiving, whilst backed up off-policy estimates should be more representative of the value function at the optimal policy, but evolve much slower.
By using both estimates, NEC can trade-off between these two estimation procedures and their relative strengths and weaknesses (speed of reward propagation vs optimality).
Finally, by having a slow changing, stable representation provided by a convolutional neural network, keys stored in the DND remain relative stable.

Our work suggests that non-parametric methods are a promising addition to the deep reinforcement learning toolbox, especially where data efficiency is paramount.
In our experiments we saw that at the beginning of learning NEC outperforms other agents in terms of learning speed. We saw that later in learning Prioritised Replay has higher performance than NEC. We leave it to future work to further improve NEC so that its long term final performance is significantly superior to parametric agents.
Another avenue of further research would be to apply the method discussed in this paper to a wider range of tasks such as visually more complex 3D worlds or real world tasks where data efficiency is of great importance due to the high cost of acquiring data.

\paragraph{Acknowledgements}
The authors would like to thank
Daniel Zoran, Dharshan Kumaran, Jane Wang,
Dan Belov,
Ruiqi Guo,
Yori Zwols,
Jack Rae,
Andreas Kirsch,
Peter Dayan, David Silver and many others at DeepMind for insightful discussions and feedback.
We also thank
Georg Ostrovski,
Tom Schaul, and
Hubert Soyer
for providing baseline learning curves.

\bibliography{bibliography}
\bibliographystyle{icml2017}
\appendix
\section{Scores on Atari Games}
\begin{table*}
\small
\begin{tabular*}{\linewidth}{rrrrrrrr}
\hline
                     &      A3C &      Nature DQN &             MFEC &              NEC &   Prioritised &   Q$^*(\lambda)$ &   Retrace$(\lambda)$ \\
                     & & & & & Replay & &\\
\hline
               Alien &    415.5 &           634.8 &           1717.7 &  \textbf{3460.6} &                800.5 &            476.8 &                541.2 \\
              Amidar &     96.3 &           126.8 &            370.9 &   \textbf{811.3} &                 99.1 &            134.5 &                162.9 \\
             Assault &    720.8 & \textbf{1489.5} &            510.2 &            599.9 &               1339.9 &           1026.6 &               1331.1 \\
             Asterix &    301.6 & \textbf{2989.1} &           1776.6 &           2480.4 &               2599.7 &           2588.6 &               2520.3 \\
           Asteroids &   1360.1 &           395.3 &  \textbf{4706.8} &           2496.1 &                854.0 &            569.8 &                579.2 \\
            Atlantis &  36383   &         14210.5 & \textbf{95499.4} &          51208.0 &              12579.1 &          28818.8 &              44771.1 \\
          Bank Heist &     15.8 &            29.3 &            163.7 &   \textbf{343.3} &                 70.1 &             32.8 &                 26.3 \\
          Battlezone &   2354.2 &          6961.0 & \textbf{19053.6} &          13345.5 &              13500.0 &           8227.2 &               6762.2 \\
           Beamrider &    450.2 & \textbf{3741.7} &            858.8 &            749.6 &               3249.6 &            656.2 &                725.4 \\
             Berzerk &    593.6 &           484.2 &   \textbf{924.2} &            852.8 &                575.6 &            647.9 &                701.5 \\
             Bowling &     25   &            35.0 &             51.8 &    \textbf{71.8} &                 30.0 &             28.4 &                 39.9 \\
              Boxing &      2.5 &            31.3 &             10.7 &    \textbf{72.8} &                 64.7 &             22.3 &                 30.7 \\
            Breakout &      1.5 &            36.8 &    \textbf{86.2} &             13.6 &                 17.7 &              6.3 &                 10.2 \\
           Centipede &   3228   &          4401.4 & \textbf{20608.8} &          12314.5 &               4694.1 &           4097.5 &               4792.9 \\
     Chopper Command &   1036.7 &           827.2 &           3075.6 &  \textbf{5070.3} &               1426.5 &            760.6 &                801.6 \\
       Crazy Climber &  70103.5 &         66061.6 &           9892.2 &          34344.0 &     \textbf{76574.1} &          64980.6 &              54177.6 \\
            Defender &   4596   &          2877.9 & \textbf{10052.8} &           6126.1 &               3486.4 &           3260.8 &               3275.6 \\
        Demon Attack &    346.8 &          5541.9 &           1081.8 &            641.4 &      \textbf{6503.6} &           4914.8 &               4836.6 \\
         Double Dunk &    -17.2 &           -19.0 &            -13.2 &     \textbf{1.8} &                -15.9 &            -18.2 &                -18.3 \\
              Enduro &      0   &           364.9 &              0.0 &              1.4 &      \textbf{1125.8} &            396.0 &                440.6 \\
       Fishing Derby &    -89.5 &           -81.6 &            -90.3 &            -72.2 &       \textbf{-48.2} &            -84.2 &                -79.8 \\
             Freeway &      0   &            21.5 &              0.6 &             13.5 &                 18.6 &    \textbf{22.2} &                 17.1 \\
           Frostbite &    218.9 &           339.1 &            925.1 &  \textbf{2747.4} &                711.3 &            407.2 &                325.0 \\
              Gopher &    854.1 &          1111.2 &  \textbf{4412.6} &           2432.3 &               1235.3 &           2292.4 &               3050.4 \\
            Gravitar &    215.8 &           154.7 &           1011.3 &  \textbf{1257.0} &                218.9 &            121.9 &                108.9 \\
            H.E.R.O. &   4598.2 &          1050.7 &          14767.7 & \textbf{16265.3} &               5164.5 &           2223.3 &               3298.2 \\
          Ice Hockey &     -8.1 &            -4.5 &             -6.5 &    \textbf{-1.6} &                -10.2 &            -11.1 &                 -9.1 \\
          James Bond &     31.5 &           165.9 &            244.7 &   \textbf{376.8} &                203.8 &             64.5 &                 67.2 \\
            Kangaroo &     55.2 &           519.6 &           2465.7 &  \textbf{2489.1} &                616.7 &            520.7 &                554.6 \\
               Krull &   3627.6 &          6015.1 &           4555.2 &           5179.2 &               6700.7 &  \textbf{8169.8} &               7399.3 \\
      Kung Fu Master &   6634.6 &         17166.1 &          12906.5 & \textbf{30568.1} &              21456.2 &          13874.7 &              18065.8 \\
 Montezuma's Revenge &      0.1 &             0.0 &    \textbf{76.4} &             42.1 &                  0.0 &              0.4 &                  2.6 \\
         Ms. Pac-Man &    770   &          1657.0 &           3802.7 &  \textbf{4142.8} &               1558.3 &           1289.9 &               1401.6 \\
      Name This Game &   2745.1 &          6380.2 &           4845.1 &           5532.0 &      \textbf{7525.0} &           5378.5 &               5227.8 \\
             Phoenix &   2542.5 &          5357.0 &           5334.5 &           5756.5 &     \textbf{11813.3} &           5771.2 &               6046.7 \\
            Pitfall! &    -43.9 &    \textbf{0.0} &            -79.0 &              0.0 &                  0.0 &             -4.4 &                 -1.5 \\
                Pong &    -20.3 &            -3.2 &            -20.0 &    \textbf{20.4} &                  6.6 &            -18.9 &                -13.3 \\
         Private Eye &     86.3 &           100.0 &  \textbf{3963.8} &            162.2 &                100.0 &           1230.4 &                 80.2 \\
              Q*bert &    438.9 &          2372.5 & \textbf{12500.4} &           7419.2 &                839.0 &           1812.4 &               2582.1 \\
          River Raid &   2312.6 &          3144.9 &           4195.0 &  \textbf{5498.1} &               4871.8 &           2787.1 &               2671.0 \\
         Road Runner &    759.9 &          7285.4 &           5432.1 &          12661.4 &     \textbf{24746.6} &           3133.1 &               6285.0 \\
          Robot Tank &      2.4 &   \textbf{14.6} &              7.3 &             11.1 &                  8.5 &             10.1 &                  9.1 \\
            Seaquest &    514.1 &           618.7 &            711.6 &           1015.3 &      \textbf{1192.2} &            611.7 &                574.3 \\
              Skiing & -20002.7 &        -19818.0 &         -15278.9 &         -26340.7 &    \textbf{-12762.4} &         -17055.7 &             -13880.4 \\
             Solaris &   2932.7 &          1343.0 &  \textbf{8717.5} &           7201.0 &               1397.1 &           2460.0 &               3211.8 \\
      Space Invaders &    201   &           642.2 &  \textbf{2027.8} &           1016.0 &                673.0 &            545.6 &                527.9 \\
          Stargunner &    613.6 &           604.8 & \textbf{14843.9} &           1171.4 &               1131.4 &            877.0 &                886.7 \\
            Surround &     -9.9 &            -9.7 &             -9.9 &    \textbf{-7.9} &                 -8.5 &             -9.8 &                 -9.9 \\
              Tennis &    -23.8 &    \textbf{0.0} &            -23.7 &             -1.8 &                  0.0 &             -4.3 &                  0.0 \\
          Time Pilot &   3683.5 &          1952.0 & \textbf{10751.3} &          10282.7 &               2430.2 &           2323.7 &               2576.0 \\
           Tutankham &    108.3 &           148.7 &             86.3 &            121.6 &       \textbf{194.0} &            108.3 &                122.4 \\
           Up'n Down &   3322.3 &         18964.9 &          22320.8 & \textbf{39823.3} &              11856.2 &          11961.2 &              13308.4 \\
             Venture &      0   &             3.8 &              0.0 &              0.0 &                  0.0 &             21.5 &        \textbf{75.6} \\
       Video Pinball &  30548.5 &         14316.0 & \textbf{90507.7} &          22842.6 &              24254.5 &          11507.3 &              14178.9 \\
       Wizard of Wor &    876   &           401.4 & \textbf{12803.1} &           8480.7 &               1146.6 &            526.8 &                420.4 \\
       Yars' Revenge &   9953   &          7614.1 &           5956.7 & \textbf{21490.5} &               9228.5 &           8884.4 &               8532.7 \\
              Zaxxon &     39.7 &           200.3 &           6288.1 & \textbf{10082.4} &               3123.5 &            278.3 &                168.3 \\
\hline
\end{tabular*}
\caption{Scores at 10 Million Frames}
\label{tab:scores-at-10}
\end{table*}
\end{document}